\documentclass[letterpaper, 10 pt, conference]{ieeeconf}  %

\IEEEoverridecommandlockouts                              %

\usepackage{graphicx}
\usepackage{subfig}
\usepackage{amssymb,amsmath,physics}
\usepackage{array,booktabs,arydshln}
\usepackage[cal=boondox,scr=boondoxo]{mathalfa}
\usepackage{algorithm}
\usepackage{algorithmic}
\usepackage[utf8]{inputenc}
\usepackage[english]{babel}
\usepackage{xcolor}
\definecolor{green}{RGB}{3,112,15}
\definecolor{yellow}{RGB}{255,140,0}
\usepackage{hyperref}
\usepackage{cite}
\usepackage{bm}
\usepackage{sidecap}

\DeclareMathOperator*{\argmin}{argmin}

\title{\Large \bf
From Agile Ground to Aerial Navigation: Learning from Learned Hallucination
}

\author{Zizhao Wang$^{1}$, Xuesu Xiao$^{2}$, Alexander J Nettekoven$^{3}$, Kadhiravan Umasankar$^{4}$,\\ Anika Singh$^{1}$, Sriram Bommakanti$^{4}$, Ufuk Topcu$^{4}$, and Peter Stone$^{2, 5}$%
\thanks{
$^{1}$Department of Electrical and Computer Engineering {\tt\small \{zizhao.wang, anikasingh\}@utexas.edu}, $^{2}$Computer Science {\tt\small \{xiao, pstone\}@cs.utexas.edu}, 
$^{3}$Mechanical Engineering {\tt\small nettekoven@utexas.edu}, 
$^{4}$Aerospace Engineering and Engineering Mechanics {\tt\small \{kadhirus99, sriram.bommakanti, utopcu\}@utexas.edu},
University of Texas at Austin, Austin, Texas 78712.
$^{5}$Sony AI.}%
\thanks{
This work has taken place in the Learning Agents Research Group (LARG) and Autonomous System Group at UT Austin, supported in part by NSF (CPS-1739964, IIS-1724157, NRI-1925082), ONR (N00014-18-2243, N00014-20-1-2115), FLI (RFP2-000), ARO (W911NF-19-2-0333), ARL (W911NF2020132), NASA (80NSSC19K0209), DARPA, Lockheed Martin, GM, and Bosch.  Peter Stone serves as the Executive Director of Sony AI America and receives financial compensation for this work.  The terms of this arrangement have been reviewed and approved by the University of Texas at Austin in accordance with its policy on objectivity in research.}%
}

\usepackage{fancyhdr}

\begin{document}
\maketitle
\thispagestyle{fancy}

\begin{abstract}
This paper presents a self-supervised Learning from Learned Hallucination (LfLH) method to learn fast and reactive motion planners for ground and aerial robots to navigate through highly constrained environments. 
The recent Learning from Hallucination (LfH) paradigm for autonomous navigation executes motion plans by random exploration in completely safe obstacle-free spaces, uses hand-crafted hallucination techniques to add imaginary obstacles to the robot's perception, and then learns motion planners to navigate in realistic, highly-constrained, dangerous spaces.  
However, current hand-crafted hallucination techniques need to be tailored for specific robot types (e.g., a differential drive ground vehicle), and use approximations heavily dependent on certain assumptions (e.g., a short planning horizon). 
In this work, instead of manually designing hallucination functions, LfLH \emph{learns} to hallucinate obstacle configurations, where the motion plans from random exploration in open space are optimal, in a self-supervised manner. LfLH is robust to different robot types and does not make assumptions about the planning horizon. 
Evaluated in both simulated and physical environments with a ground and an aerial robot, LfLH outperforms or performs comparably to previous hallucination approaches, along with sampling- and optimization-based classical methods. 
\end{abstract}

\section{INTRODUCTION}
\label{sec::intro}

Although classical navigation systems can safely and reliably move mobile robots from one point to another within obstacle-occupied environments, recent machine learning techniques have demonstrated improvement over their classical counterparts~\cite{xiao2020motion}, e.g., by learning local planners~\cite{liu2021lifelong, xiao2021learning}, learning world representation~\cite{wigness2018robot, richter2017safe}, or learning planner parameters~\cite{xiao2020appld, wang2021appli, xu2021applr}. However, these learning approaches heavily depend on access to high quality training data. 

Learning from Hallucination (LfH)~\cite{xiao2021toward, xiao2021agile} is a recently proposed paradigm to address the difficulty of 1) obtaining high-quality training data for traditional Imitation Learning (IL) from expert demonstrations~\cite{pfeiffer2017perception} and 2) Reinforcement Learning (RL) from trial-and-error~\cite{chiang2019learning}. During LfH training, the robot executes a variety of random motion plans in a completely safe open space, imagines obstacle configurations for which the motion plans are optimal (called \emph{hallucination}), and learns an end-to-end local planner as a mapping from the hallucinated obstacle configurations to the optimal motion plans in open space. The inherent safety of navigating in a completely open training environment allows generation of a large amount of training data with no expert supervision or costly failures during trial-and-error learning. During deployment, learned local planners react to real obstacles within constant time (i.e., querying a pre-trained neural network), which is not dependent on how densely packed the surrounding obstacles are. 

However, existing LfH methods require manually designed hallucination functions to generate the most constrained~\cite{xiao2021toward} or a minimal~\cite{xiao2021agile} obstacle set. While the former requires access to a fine-resolution global path and runtime hallucination during deployment, the latter assumes a short planning horizon to assure that one representative minimal unreachable set can approximately represent all possible minimal unreachable sets. Furthermore, these limitations prevent the manually-designed hallucination functions from extending from simple differential-drive ground vehicles to different robot types, e.g., aerial robots. 

The Learning from Learned Hallucination (LfLH) method introduced in this work removes the necessity of manually designing hallucination functions specific to certain robot types and assumptions. In a self-supervised manner, LfLH automatically learns distributions of obstacles which make randomly explored motion plans in open space optimal, samples obstacle configurations from learned obstacle distributions, and finally learns a local planner that maps hallucinated obstacles to optimal motion plans. During deployment, the robot precepts real obstacles and uses the learned local planner to generate motion plans.
LfLH is tested on a ground and an aerial robot, both in simulated benchmark testbeds~\cite{perille2020benchmarking} and physical environments. Superior navigation performance is achieved compared to existing LfH approaches~\cite{xiao2021toward, xiao2021agile} and classical sampling-based~\cite{fox1997dynamic} and optimization-based~\cite{zhou2020ego} planners. 

\section{RELATED WORK}
\label{sec::related}
This section reviews classical motion planning and recent machine learning techniques for mobile robot navigation. 

\subsection{Classical Motion Planning}
Classical motion planning techniques for mobile robot navigation mostly work in the robot Configuration Space (C-Space)~\cite{lavalle2006planning} and mainly comprise two categories: sampling-based and optimization-based. Sampling-based motion planners~\cite{fox1997dynamic} generate sample motion plans and select the best sample based on a certain metric, such as maximum clearance, shortest path, or a combination thereof~\cite{lavalle2006planning}. Optimization-based planners~\cite{zhou2020ego} start with an initial motion plan, then use optimization techniques to iteratively refine the initial plan to avoid obstacles while observing kinodynamic constraints. One common shortcoming of both categories is that when dealing with more constrained obstacle spaces, classical motion planners require increased computation: sampling-based planners require more samples to find a collision-free motion plan to go through all obstacles, while optimization-based planners require more optimization iterations until a feasible plan can satisfy both collision and kinodynamic constraints. 

Compared with classical motion planning algorithms, one advantage of the proposed LfLH approach is that its computation is not dependent on obstacle density during deployment, because LfLH simply queries a pre-trained neural network to produce feasible and fast navigation behaviors. 

\subsection{Machine Learning for Navigation}
Machine learning approaches have been applied to the classical navigation pipeline in different ways~\cite{xiao2020motion}, such as constructing a world representation~\cite{wigness2018robot, richter2017safe}, fine-tuning planner parameters~\cite{xiao2020appld, wang2021appli, xu2021applr}, improving navigation performance with experience~\cite{liu2021lifelong}, or enabling social~\cite{chen2017socially} and terrain-aware navigation~\cite{xiao2021learning}. Most learning methods require either extensive (RL) or high-quality (IL) training data, such as that derived from trial-and-error exploration or from human demonstrations, respectively. 

LfH~\cite{xiao2021toward, xiao2021agile} has been recently proposed to alleviate the difficulty of acquiring extensive or high-quality training data: from random exploration in a completely safe open space with complete safety, motion planners can be learned by synthetically projecting the \emph{most constrained}~\cite{xiao2021toward} or augmented \emph{minimal}~\cite{xiao2021agile} C-space onto the robot perception. Through carefully designed hallucination functions, these methods have shown fast and agile maneuvers on ground robots compared to classical motion planning and traditional learning approaches. However, the design of specific hallucination functions does not easily extend to other robot types (e.g., aerial robots~\cite{xiao2021toward}) and relies on specific assumptions (e.g., a short motion plan/planning horizon to make approximated hallucination valid~\cite{xiao2021agile}). 

LfLH removes the requirement for a carefully designed hallucination function tailored to a specific robot with strict assumptions, and instead \emph{learns} hallucinated obstacle distributions which assure the motion plans executed in open space are optimal in a self-supervised manner. Sample obstacle configurations can be drawn from the learned obstacle distributions as training data to learn a motion planner.

\section{APPROACH}
\label{sec::approach}
\begin{figure*}
  \centering
  \includegraphics[width=1.7\columnwidth]{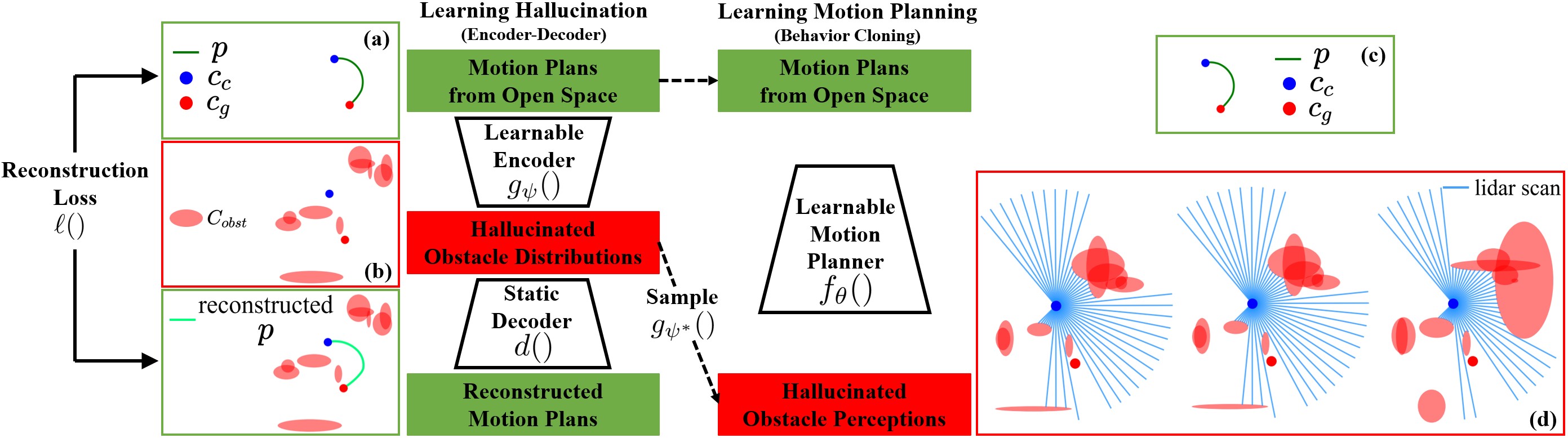}
\caption{\small The Encoder-Decoder architecture learns hallucination function $g_{\psi} $ from motion plans in open space (a) to obstacle distributions (b). Sampling from hallucinated obstacle distributions and rendering corresponding observations (d), motion planner $f_{\theta}$ is learned with Behavior Cloning using motion plans collected in the open space as ground truth (c). During deployment, $f_{\theta}$ takes in real obstacle perceptions and generates corresponding motion plans.}
  \label{fig::diagram}
  \vspace{-16pt}
\end{figure*}

In this section, we present our Learning from Learned Hallucination (LfLH) approach. We first formulate the problem using the LfH framework, then present the proposed approach to \emph{learn} (instead of manually design) a hallucination function, from which a motion planner is finally learned, as shown in Fig. \ref{fig::diagram}.

\subsection{Problem Definition}
We adopt the same notation used by Xiao {\em et al}. to formalize LfH~\cite{xiao2021toward} and Hallucinated Learning and Sober Deployment (HLSD)~\cite{xiao2021agile}: given a robot's C-space partitioned by unreachable (obstacle) and reachable (free) configurations, $C = C_{obst} \cup C_{free}$, 
the classical motion planning problem is to find a function $f(\cdot)$ that can be used to produce optimal plans $p=f(C_{obst}~|~c_c, c_g)$ that result in the robot moving from the robot's current configuration $c_c$ to a specified goal configuration $c_g$ without intersecting (the interior of) $C_{obst}$. Here, a plan $p \in \mathcal{P}$ comprises a sequence of actions $\{u_i\}_{i=1}^{t}$ ($u_i \in \mathcal{U}$, $\mathcal{P}$ and $\mathcal{U}$ are the robot’s plan and action space, respectively). Considering the inverse problem of finding $f(\cdot)$, LfH~\cite{xiao2021toward} and HLSD~\cite{xiao2021agile} use hallucination functions denoted as $g(p~|~c_c, c_g)$, to generate the (unique) \emph{most constrained} and a (not unique) \emph{minimal} obstacle set, respectively, to make a motion plan $p$ generated by a random policy $\pi_{rand}$ in open space optimal. To instantiate these hallucination functions, hand-crafted rules are designed for specific robot types (e.g. differential drive robots) and do not easily extend to others. LfH~\cite{xiao2021toward} further requires a fine-resolution global path and a runtime hallucination function $h(\cdot)$ to augment the real obstacle perceptions to the most constrained cases during deployment. HLSD~\cite{xiao2021agile} uses one representative minimal unreachable set to approximately represent all of them. This approximation is accurate only for short planning horizons or motion plans, either requiring frequent replanning or limiting navigation speed at runtime. 

LfLH aims to learn a parameterized hallucination function $g_\psi(\cdot)$, which outputs probability distributions of obstacles, in a self-supervised manner, without the need to manually design hallucination functions for each robot type and thus avoiding the subsequent problems described above. Then it samples many times from this learned distribution, $C_{obst} \sim g_\psi(\cdot)$, to generate many obstacle configurations, in which the free-exploration motion plans in open space are close to optimal. 

\subsection{Learning Hallucination}
We adopt an encoder-decoder architecture to learn the hallucination function $g_\psi(p~|~c_c, c_g)$ parameterized by $\psi$. 
Taking the current configuration $c_c$, goal configuration $c_g$, and the corresponding plan $p$ as input, the encoder $g_\psi(\cdot)$ generates probability distributions of obstacles. We assume obstacles are ellipses (or ellipsoids), and thus the obstacle distributions are modeled as normal distributions of obstacle locations and sizes in the C-Space.
To shape the obstacle distributions such that the given plan $p$ is optimal, LfLH uses a classical motion planner which does not have learnable parameters as the decoder $d(C_{obst} \sim g_\psi(p~|~c_c, c_g))$. The decoder $d(\cdot)$ samples from the obstacle distributions and then computes the optimal motion plan for the sampled obstacles using its built-in algorithms. If the reconstructed motion plan is the same as the given plan $p$, it indicates $p$ is also optimal for the sampled obstacles.
In this work, we choose $d(\cdot)$ to be a differentiable optimization-based motion planner for easy training. To use non-differentiable motion planners, one can use approximate gradients algorithms, like REINFORCE \cite{williams1992simple}. 
Overall, our encoder-decoder architecture uses gradient descent to find the optimal parameters $\psi^*$ for $g_\psi(\cdot)$ by minimizing a self-supervised loss using a dataset $P$ of motion plans collected using random exploration in an open space: 
\begin{equation}
\psi^* = \argmin_\psi \mathop{\mathbb{E}}_{\substack{p \sim P\\C_{obst} \sim g_\psi(p|c_c, c_g)}} \big[ \ell(p, d(C_{obst})) + \ell_r(C_{obst}, p)\big].
\label{eqn::psi}
\end{equation}
The choice of the reconstruction loss function $\ell(\cdot, \cdot)$ depends on the representation of a motion plan $p$ (e.g., a sequence of raw motor commands or a B-spline trajectory). In addition to the main reconstruction loss $\ell(\cdot, \cdot)$, a regulization loss $\ell_r$ is employed to stablize training, 
\begin{equation}
\ell_r(C_{obst}, p) = \lambda_1 \ell_{prior}(C_{obst}) +  \lambda_2 \ell_{coll}(C_{obst},p), 
\label{eqn::regularization}
\end{equation}
where $\ell_{prior}$ prevents $g_\psi$ from overfitting by encouraging a larger probability that $C_{obst}$ is sampled from a prior distribution (i.e., minimizing the discrepancy between the output of $g_\psi$ and the prior), $\ell_{coll}$ is the penalty for collision between obstacles $C_{obst}$ and plan $p$ as well as between each pair of obstacles in $C_{obst}$, and $\lambda_1, \lambda_2$ are corresponding regularization weights. Implementation details of $\ell$, $\ell_{prior}$, and $\ell_{coll}$ can be found in Sec. \ref{sec::experiments}.

After learning $g_{\psi^*}(\cdot)$ in a self-supervised manner, we can sample from the learned obstacle distributions $g_{\psi^*}(p~|~c_c, c_g)$ many times to generate many obstacle configurations $C_{obst}$ where the motion plan $p$ is close to optimal. We then form a training set $\mathcal{D}_{train}$ with individual data points $(C_{obst}, p, c_c, c_g)$. The specific instantiation of $C_{obst}$ depends on the perception modality of the mobile robot. For example, for ground robots with 2D LiDAR, we use 2D ray casting from the onboard sensor to the sampled obstacles to determine range readings for each laser beam; for aerial robots with 3D depth cameras, we use 3D rendering to determine each camera pixel's depth value to the sampled obstacles. 

\subsection{Learning from Learned Hallucination}
With the training set $\mathcal{D}_{train}$ constructed using the learned hallucination function $g_{\psi^*}(\cdot)$, we learn a parameterized motion planner $f_\theta(\cdot)$ from $\mathcal{D}_{train}$ by minimizing a supervised learning loss using gradient descent: 
\begin{equation}
\theta^* = \argmin_\theta \mathop{\mathbb{E}}_{(C_{obst}, p, c_c, c_g) \sim \mathcal{D}_{train}} \big[ \ell(p, f_\theta(C_{obst}~|~c_c, c_g))\big].
\label{eqn::theta}
\end{equation}
The loss function $\ell(\cdot, \cdot)$ in Eqns. \ref{eqn::psi} and \ref{eqn::theta} could take the same form. But in practice, the encoder $g_{\psi}(\cdot)$'s input plan (Eqn. \ref{eqn::psi}) and the motion planner $f_{\theta}(\cdot)$'s output plan (Eqn. \ref{eqn::theta}) can take different forms to faciliate learning. Details can be found in Sec. \ref{sec::experiments}.
During deployment, the learned $f_{\theta^*}$ uses real obstacle perceptions rather than hallucinated ones to generate effective motion plans.

The entire learning pipeline is shown in Alg. \ref{alg::lflh}, including data collection, learning hallucination, learning from learned hallucination, and deployment. 

\begin{algorithm}[b!]
 \caption{Learning from Learned Hallucination}
 \begin{algorithmic}[1]
 \renewcommand{\algorithmicrequire}{\textbf{Input:}}
 \REQUIRE $\pi_{rand}$, $g_{\psi}(\cdot)$, \emph{sampling count}, $f_{\theta}(\cdot)$
\\\hrulefill
  \STATE // \textbf{Data Collection}
  \STATE collect motion plans $(p, c_c, c_g)$ in free space with $\pi_{rand}$ to form motion plan dataset $P$ 
  \STATE // \textbf{Learning Hallucination}
  \STATE learn $\psi^*$ using Eqn. \ref{eqn::psi} for $g_\psi(\cdot)$ with $P$
  \STATE $\mathcal{D}_{train} \leftarrow \emptyset$
  \FOR {every $(p, c_c, c_g)$ in $P$}
    \FOR {\emph{sampling count} times}
        \STATE sample $C_{obst} \sim g_{\psi^*}(p~|~c_c, c_g)$
        \STATE $\mathcal{D}_{train}=\mathcal{D}_{train}\cup(C_{obst}, p, c_c, c_g)$
    \ENDFOR
  \ENDFOR
  \STATE // \textbf{Learning from Learned Hallucination}
  \STATE learn $\theta^*$ using Eqn. \ref{eqn::theta} for $f_\theta(\cdot)$ with $\mathcal{D}_{train}$
\\\hrulefill
  \STATE // \textbf{Deployment} (each time step)
  \STATE receive $C_{obst}, c_c, c_g$
  \STATE plan $p = \{u_i\}_{i=1}^{t} = f_{\theta^*}(C_{obst}~|~c_c, c_g)$
 \RETURN $p$
 \end{algorithmic}
 \label{alg::lflh}
 \end{algorithm}

\section{EXPERIMENTS}
\label{sec::experiments}
LfLH is implemented on a ground and an aerial robot to validate our hypothesis that LfLH can automatically learn obstacle configurations where motion plans executed in open space are near-optimal, and agile motion planners can be learned through the learned halluciantion. 

\subsection{Ground Robot}
We first implement LfLH on a ground robot and compare LfLH's performance with a classical sampling-based motion planner~\cite{fox1997dynamic} and state-of-the-art learning approaches from hallucination, including LfH~\cite{xiao2021toward} and HLSD~\cite{xiao2021agile}. 

\subsubsection{Implementation}
\label{subsec::ugv_implementation}
We use a Clearpath Jackal robot, a four-wheeled, differential-drive, Unmanned Ground Vehicle (UGV), running the Robot Operating System \texttt{move\textunderscore base} navigation stack. The Jackal has a 720-dimensional 2D LiDAR with a 270$^\circ$ field of view, which is used to instantiate obstacle configuration $C_{obst}$ using ray casting. Its DWA local planner is replaced with LfLH. 

For data collection (line 2 in Alg. \ref{alg::lflh}), we collect three separate datasets: a dataset with mostly constant 0.4m/s linear velocity ($v$) and varying angular velocity ($\omega$); a varying $v$ and $\omega$ dataset with 1.0m/s max $v$; and a varying $v$ and $\omega$ dataset with 2.0m/s max $v$. 
To be specific, $\pi_{rand}$ samples a target $\left(\hat{v}, \hat{\omega}\right)$ pair at random according to the max limit and commands the UGV to reach that speed with constant increments/decrements considering its acceleration limit. 
The most constrained hallucination in LfH~\cite{xiao2021toward} only works with the 0.4m/s dataset, while being confused by the ambiguity of varying $v$ in 1.0m/s and 2.0m/s datasets in the most constrained spaces. The minimal hallucination in HLSD~\cite{xiao2021agile} works with both the 0.4m/s and 1.0m/s datasets, but does not learn well from the 2.0m/s dataset because the approximation of the representative minimal unreachable set used in HLSD fails to represent all minimal sets in the long trajectory produced by the faster speed. Only LfLH works well with all datasets. 

For the UGV, a single odometry point $o_i$ consists of position $x_i$, orientation $\phi_i$, linear velocity $v_i$, and angular velocity $\omega_i$. We use $N=125$ odometry points to compose a motion plan $p$ that takes $2.5$s and train the hallucination function $g_{\psi^*}(\cdot)$ with Eqn. \ref{eqn::psi} (line 4 in Alg. \ref{alg::lflh}). The encoder $g_{\psi}(\cdot)$ is modeled as a network with three one-dimensional convolutional layers to extract temporal features and a fully-connected layer to map temporal features to ten obstacles' location and size distributions, in the form of means and variances. The decoder $d$ is Ego-Planner \cite{zhou2020ego}, an optimization-based planning algorithm which we reimplement with differential convex optimization layers \cite{agrawal2019differentiable} to enable differentiation through $d$ itself.
The reconstruction loss $\ell$ in Eqn. \ref{eqn::psi} is the mean squared error of all positions and linear velocities $\{x_i, v_i\}_{i=1}^{N}$ in $p$ and their reconstructed values.
For regularization, the obstacle location prior distribution is a normal distribution fitted on all positions $\{x_i\}_{i=1}^{N}$ in the plan $p$, to prevent obstacles from getting too far away from the plan $p$. Meanwhile, the obstacle size prior is a normal distribution with an empirically chosen mean of 0.3m and variance of 0.0025m$^2$. The obstacle-obstacle/obstacle-plan collision regularization loss is $\ell_{coll} = \sum\max(c-d,0)^2$ where clearance $c=$0.5m and $d$ is the distance either between two obstacles or between the obstacle and its closet point on the plan $p$. Regularization weights are tuned with grid search as $\lambda_1=0.3$ and $\lambda_2=2.0$.

After training, we construct the training set $\mathcal{D}_{train}$ by sampling ten obstacle configurations from $g_{\psi^*}(\cdot)$ for each collected plan. Beyond the ten hallucinated obstacles, to increase the variance of training samples, we randomly sample five additional obstacles whose distances to $p$ are proportional to the velocity to account for motion uncertainty at fast speeds. To make sure the training samples are valid, we filter out invalid samples where obstacles are too close to the path $p$ (clearance $c < 0.5$m). Finally, the observations, i.e., 2D Lidar scans, are rendered using ray casting given the UGV's configuration $c_c$ and obstacle configurations $C_{obst}$. The motion planner $f_{\theta^*}$ is trained to produce only the first action $(v_1, \omega_1)$ in the entire motion plan $p$ for simplicity (line 13 in Alg. \ref{alg::lflh}), and it is represented as a fully-connected network with 2 hidden layers of 256 units.
During deployment, the UGV reasons in the robot frame, so $c_c$ is the origin and $c_g$ is a point 1.5m away from the robot on the global path (line 15 in Alg. \ref{alg::lflh}). We use the same Model Predictive Control (MPC) model as HLSD \cite{xiao2021agile} to check for and avoid collisions.

\subsubsection{Simulated Experiments}
\begin{figure}
  \centering
  \includegraphics[width=1\columnwidth]{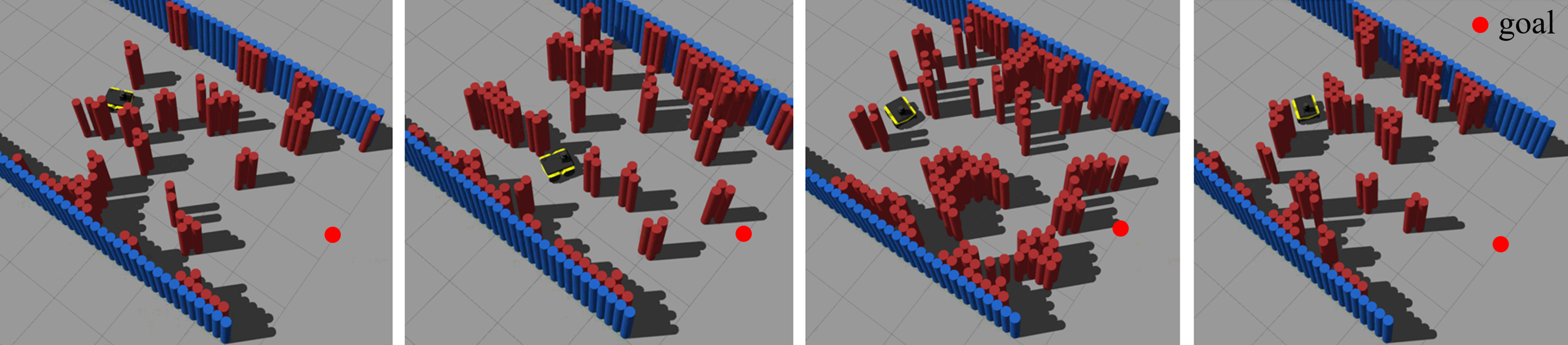}
  \vspace{-15pt}
  \caption{\small Four example BARN environments}
  \label{fig::simulated_envs}
  \vspace{-10pt}
\end{figure}

We first use the BARN dataset~\cite{perille2020benchmarking} with 300 navigation environments (example environments shown in Fig. \ref{fig::simulated_envs}) randomly generated by Cellular Automata to compare the different motion planners. As a classical sampling-based planner, DWA's~\cite{fox1997dynamic} max linear velocity is increased from the default 0.5m/s to 2.0m/s for a fair comparison with other planners. We find that by also quadrupling DWA's default sampling rate for linear and angular velocity (to 24 and 80), the UGV's performance is roughly the same as when using the default parameters (but at quadruple the speed). We train nine different planners with LfH~\cite{xiao2021toward}, HLSD~\cite{xiao2021agile}, and LfLH on all three datasets. For all the planners, we run three navigation trials in each of the 300 navigation environments in BARN between a specified start and goal location without a map and record the traversal time (with 50s maximum). 

\begin{figure}
  \centering
  \includegraphics[width=0.9\columnwidth]{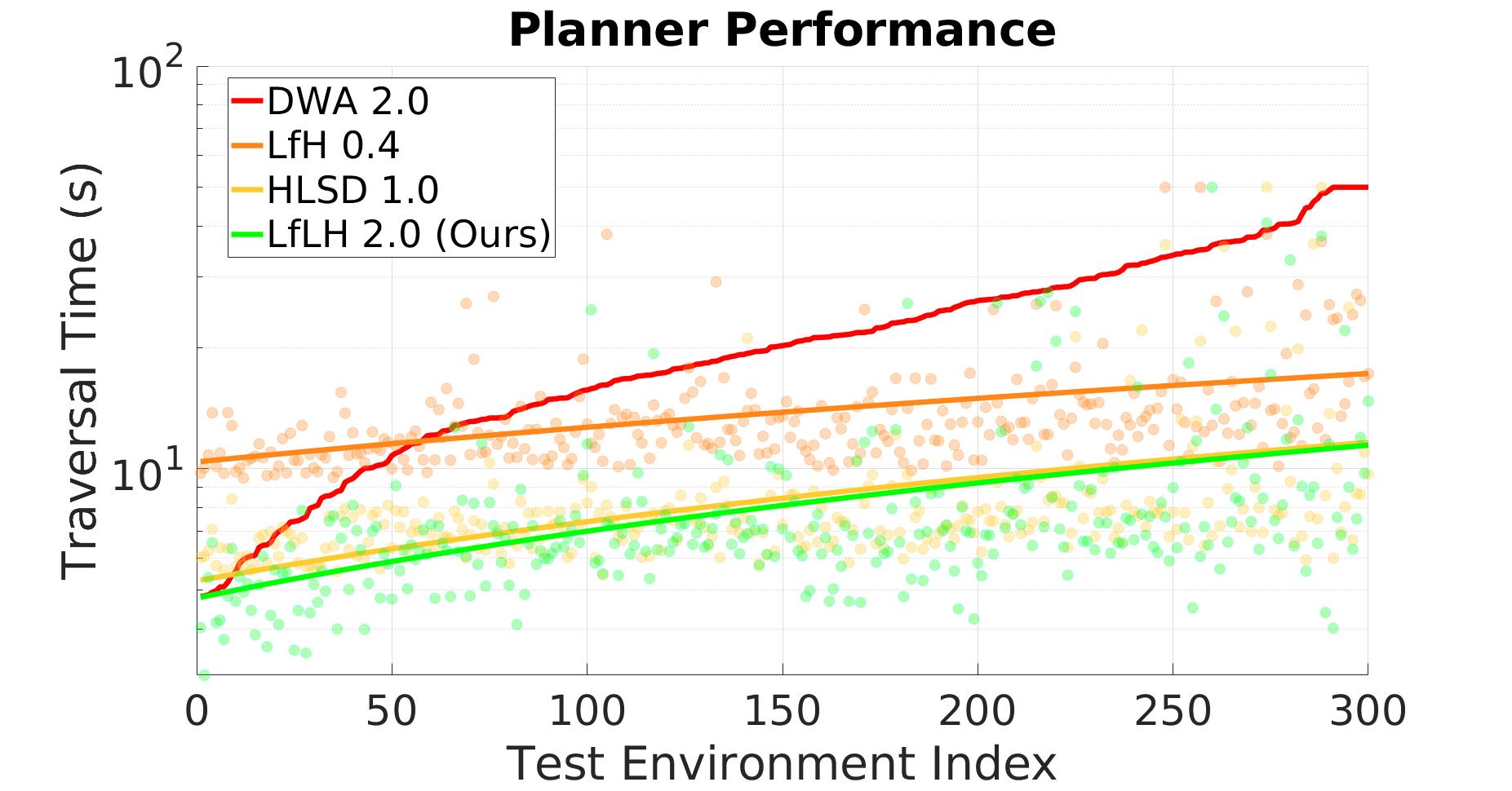}
  \caption{Simulation results in BARN}
  \label{fig::lflh_barn}
  \vspace{-10pt}
\end{figure}

We list the average traversal time with the standard deviation in Tab. \ref{tab::simulated_results}. The $\infty$ sign indicates the motion planner learned using the corresponding method and dataset fails to navigate to the goal in most of the trials (getting stuck or colliding). For the other entries, a low traversal time means efficient and agile navigation performance. LfLH is the only learning method that works well with all three datasets, and achieves the best navigation performance. 
For clarity, we only plot the test results in each environment (averaged over three trials) using each variant with its fastest working dataset, i.e., LfH 0.4, HLSD 1.0, and LfLH 2.0, along with DWA 2.0 in Fig. \ref{fig::lflh_barn} (scattered dots). We also fit a line to show the trend of each method. In Fig. \ref{fig::lflh_barn} we order the navigation environments from left to right with increasing average traversal time achieved by DWA 2.0 (red). Note that the y axis uses a log scale to better visualize the differences in the small traversal time range. LfLH 2.0 (green) achieves similar results in easy environments (left) and significantly outperforms DWA 2.0 in difficult ones (right). Although LfH 0.4 (orange) and HLSD 1.0 (yellow) have a disadvantage due to slow max speed in easy environments, they exhibit good agility in more constrained difficult ones, especially HLSD 1.0. LfLH 2.0 is the only planner that achieves good performance in both easy and difficult environments, shown by the flat green line. Note that the maximal velocity of the UGV is 2.0m/s and the highly constrained BARN environments require slow speeds for agile maneuvering in most places.

\begin{table}
\centering
\caption{UGV Simulated Average Traversal Time}
\begin{tabular}{c|cccc}
\toprule
DWA 2.0 & Dataset & LfH & HLSD & \textbf{LfLH} \\ 
\midrule
 & 0.4m/s & 13.8$\pm$5.3s  & 13.2$\pm$7.9s & 13.4$\pm$6.4s\\
22.1$\pm$11.4s & 1.0m/s & $\infty$ & 8.5$\pm$5.2s & 8.3$\pm$3.8s\\
 & 2.0m/s & $\infty$  & $\infty$ & \textbf{8.1}$\pm$5.4s\\
\bottomrule
\end{tabular}
\vspace{-10pt}
\label{tab::simulated_results}
\end{table}

\subsubsection{Physical Experiments}
\begin{figure}
  \centering
  \includegraphics[width=0.9\columnwidth]{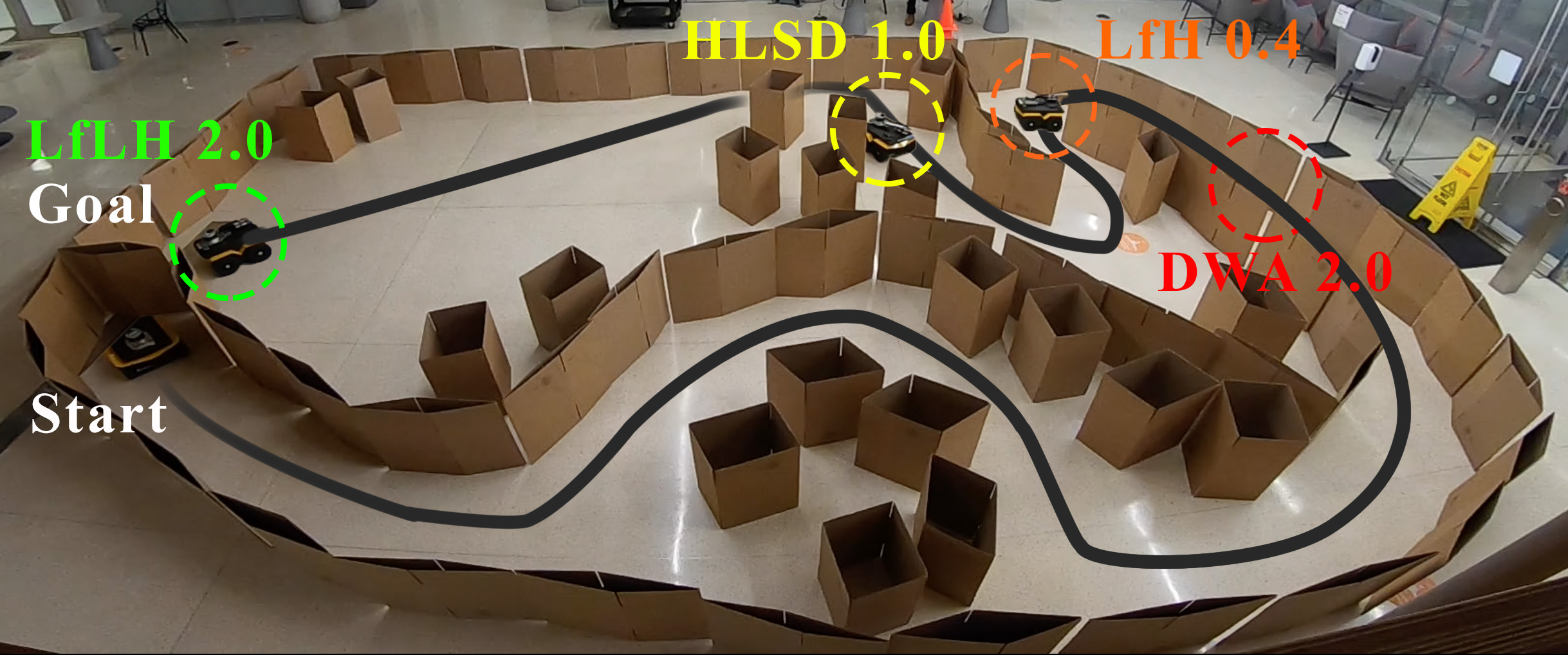}
  \vspace{-5pt}
  \caption{\small UGV Physical Experiments: Progress of HLSD 1.0 (yellow), LfH 0.4 (orange), and DWA 2.0 (red) in the obstacle course, when LfLH 2.0 (green) reaches the goal. }
  \label{fig::physical}
  \vspace{-10pt}
\end{figure}

We also deploy the four planners in a physical test course, for five trials each (Fig. \ref{fig::physical}). The results are shown in Tab. \ref{tab::physical_results}. 
The sampling-based DWA planner~\cite{fox1997dynamic} fails to sample feasible motions in many constrained spaces, and has to execute many recovery behaviors before re-sampling. Therefore DWA takes a long average time with large variance to traverse the course. LfH 0.4 requires a fine-resolution global path~\cite{xiao2021toward} and drives smoothly but slowly everywhere along the course. Being too conservative in wide open spaces causes LfH 0.4 to achieve similar results to DWA 2.0, which gets stuck in many places but makes up the time by accelerating in open spaces. 
HLSD 1.0 also navigates smoothly, but much faster than LfH 0.4, and outperforms DWA 2.0. Our LfLH 2.0 is the only planner that can learn from the fast 2.0m/s dataset, achieving the best performance among all the variants. 

\begin{table}
\centering
\caption{UGV Physical Average Traversal Time}
\begin{tabular}{cccc}
\toprule
DWA 2.0 & LfH 0.4 & HLSD 1.0 & \textbf{LfLH 2.0} \\ 
\midrule
73.6$\pm$3.8s & 78.4$\pm$1.8s  & 50.6$\pm$0.8s & \textbf{41.1}$\pm$0.9s\\
\bottomrule
\end{tabular}
\label{tab::physical_results}
\end{table}

\begin{figure}
  \centering
  \includegraphics[width=0.5\columnwidth]{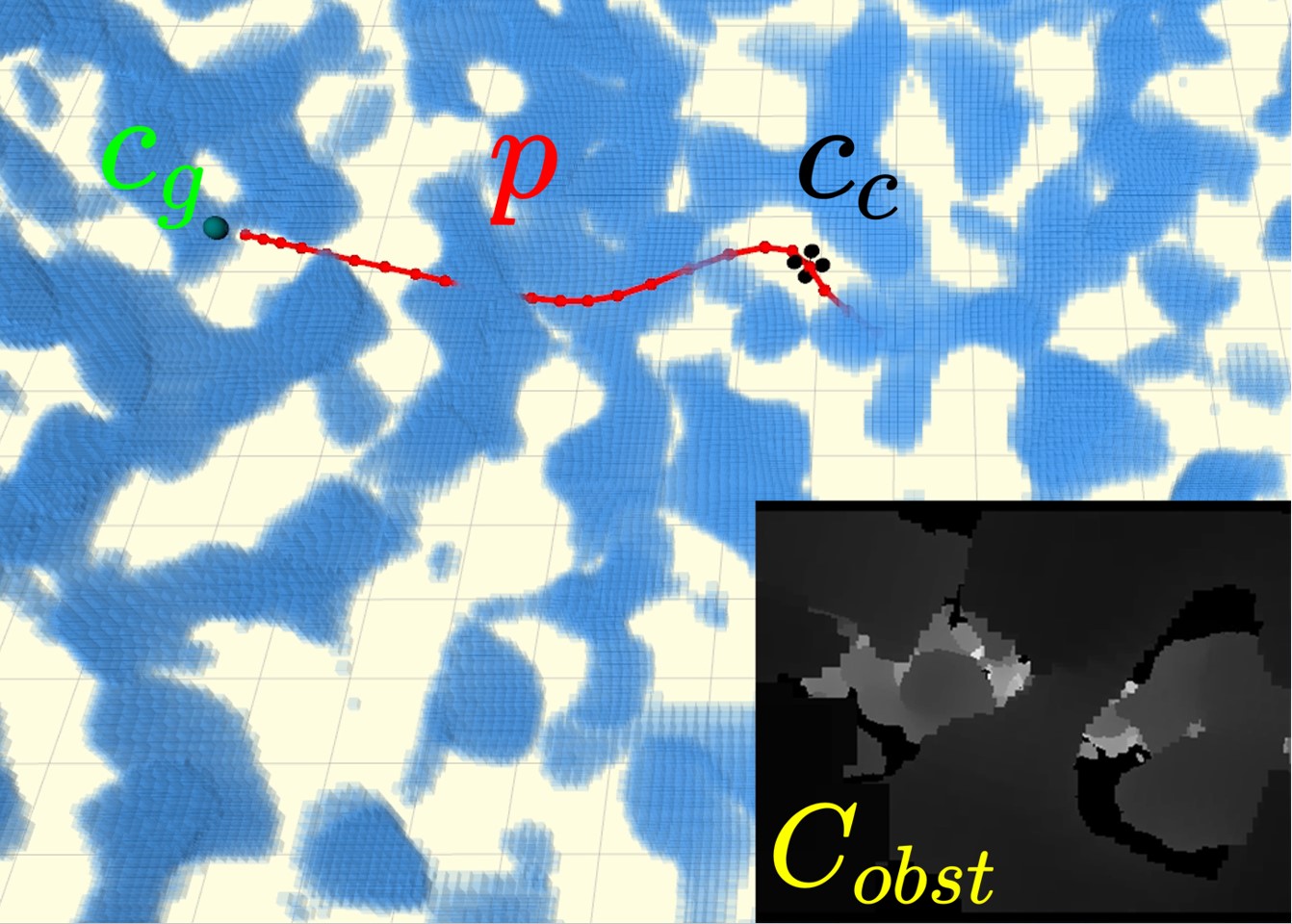}
  \vspace{-5pt}
  \caption{\small \textbf{Left}: In Ego-Planner's simulator, the obstacles are shown as blue clouds, the goal is marked green ($c_g$), and the produced motion plan $p$ is marked red. The bottom right is the depth image as an instantiation of $C_{obst}$.
  }
  \label{fig::drone_simulator}
  \vspace{-10pt}
\end{figure}

\subsection{Aerial Robot}

We also evaluate LfLH on an Unmanned Aerial Vehicle (UAV, a quadrotor) where hallucination is more challenging due to higher dimensionality and more agility. We compare LfLH with a state-of-the-art optimization-based UAV trajectory planning algorithm, Ego-Planner \cite{zhou2020ego}.

\subsubsection{Implementation}
We apply LfLH to a simulated UAV in Ego-Planner's simulator and a physical PX4 Vision UAV platform shown in Fig. \ref{fig::drone_simulator}. Both the simulated and physical UAVs use depth input, acquired by rendering in simulation and a Structure Core camera,
respectively, to instantiate obstacle configuration $C_{obst}$.

Due to unavailability of a motion capture system, we collect a dataset of 20-min flight in simulation (line 2 in Alg. \ref{alg::lflh}). We represent the random exploration policy $\pi_{rand}$ to collect $p$ in an open space using Ego-Planner with 2.0m/s max $v$ in a highly constrained environment. We only record motion plans $p$ but do not record any perception input (as if the UAV were flying in an open space),\footnote{We leave a true random exploration policy in an open space as future work, such as random teleoperation or a random policy as in the UGV case. } which is later synthesized by LfLH. Note that previous learning approaches from hallucination cannot solve this 3D aerial navigation task: LfH~\cite{xiao2021toward} is not applicable because no global path is available; The approximated minimal unreachable set in HLSD~\cite{xiao2021agile} cannot effectively represent 3D obstacles given the long flight plan.

For our LfLH implementation in 3D, the odometry point contains the same information $(x_i, \phi_i, v_i, \omega_i)$ as the UGV (but in 3D), and each plan $p$ consists of $M=500$ odometry points that takes $2.5$s. The encoder $g_\psi$, decoder $d$, reconstruction loss $\ell$, and regularization $\ell_r$ are in the same format as for the UGV, except that $g_\psi(\cdot)$ outputs distributions of 15 obstacles, the obstacle size prior has a different mean of 0.6m and variance of 0.2$m^2$ in the prior regularization loss $\ell_{prior}$, the clearance increases to $c=0.6$m in the collision regularization loss $\ell_{collsion}$, and $\lambda_1=0.5, \lambda_2=5.0$ for regularization weights.

After training $g_\psi(\cdot)$ (line 4 in Alg. \ref{alg::lflh}), the training set $\mathcal{D}_{train}$ is constructed with the collected plans $p$ and their corresponding observed $C_{obst}$, i.e., depth images rendered using ray casting given the 15 sampled hallucinated obstacles and five additional obstacles sampled in the same way as the UGV in Sec. \ref{subsec::ugv_implementation}. The motion planer $f_\theta$ is modeled as four convolutional layers and three fully-connected layers used to extract features from the depth image, goal configuration, and current velocities $(h, c_g, v, \omega)$. Then $f_\theta$ is trained to produce positions and linear velocities $\{\hat{x}_i, \hat{v}_i\}_{i=1}^M$ in the entire motion plan $p$. %
A similar MPC method to HLSD~\cite{xiao2021agile} used for ground navigation is also applied to check for collisions by the UAV.

\begin{table}
\centering
\caption{UAV Simulated Evaluation Results}
\begin{tabular}{lcc}
\toprule
Metrics & Ego-Planner & LfLH \\ 
\midrule
Survival Time (s)       & 101.99$\pm$62.83  & \textbf{192.87}$\pm$161.37 \\
Survival Distance (m)   & 174.15$\pm$106.74 & \textbf{213.07}$\pm$172.98 \\
SPL                     & \textbf{0.74}              & 0.56 \\
\bottomrule
\end{tabular}
\label{tab::drone_simulation_results}
\end{table}

\subsubsection{Simulated Environments}

We first evaluate LfH and Ego-Planner in simulation with a randomly generated forest shown in Fig. \ref{fig::drone_simulator} left. Each method is tested for 10 trials, and in each trial the UAV keeps navigating to randomly generated goals until the UAV collides with obstacles. Meanwhile, we record the total traversal time and distance, as well as the individual traversal time $t_i$ and distance $p_i$ to each goal $i$, from which we measure Success weighted by Path Length~\cite{anderson2018evaluation}, $\text{SPL} = \frac{1}{K}\sum_{i=1}^K S_i\frac{l_i}{p_i}$,
where $S_i$ is the binary indicator of success for reaching the $i$th goal, and $l_i$ is the Euclidean distance from the $i$th start to the $i$th goal, which is always smaller than $p_i$.

We list the average and standard deviation (if applicable) of these metrics in Tab. \ref{tab::drone_simulation_results}. Compared with Ego-Planner, LfLH survives (keeps navigating in a collision-free manner) longer both in terms of traversal time and distance, but it has lower SPL. In other words, LfLH trades off aggressive motions for safety. With safer motion plans, it can react faster to unexpected obstacles in the highly constrained environments (e.g., obstacles that are occluded until the robot gets pretty close). To the best of our knowledge, LfLH is the first learning-based planner for aerial robots that can navigate in such highly constrained spaces.

\section{CONCLUSIONS}
\label{sec::conclusions}

LfLH is a self-supervised machine learning technique for mobile robot navigation that only requires data in open space. In addition to self-supervised learning of a motion planner, LfLH also \emph{learns} self-supervised to generate hallucinated obstacle configurations, from which the motion planner is learned, instead of requring hand-crafted hallucination functions. In contrast to the manually designed ones ~\cite{xiao2021toward, xiao2021agile}, LfLH is robust to different robot types and can be applied to agile ground and aerial robots navigating at faster speeds. 
Although in our experiments the ground robot learns from a real random exploration policy, one future research direction is to investigate truly random exploration for aerial vehicles, instead of using trajectories collected from an existing motion planner. In addition to a physical demonstration, other interesting directions include extending LfLH to dynamic obstacles and designing better exploration strategies to cover necessary navigation skills for all possible obstacle configurations.

\bibliographystyle{IEEEtran}
\bibliography{IEEEabrv,references}

\begin{thebibliography}{10}
\providecommand{\url}[1]{#1}
\csname url@samestyle\endcsname
\providecommand{\newblock}{\relax}
\providecommand{\bibinfo}[2]{#2}
\providecommand{\BIBentrySTDinterwordspacing}{\spaceskip=0pt\relax}
\providecommand{\BIBentryALTinterwordstretchfactor}{4}
\providecommand{\BIBentryALTinterwordspacing}{\spaceskip=\fontdimen2\font plus
\BIBentryALTinterwordstretchfactor\fontdimen3\font minus
  \fontdimen4\font\relax}
\providecommand{\BIBforeignlanguage}[2]{{%
\expandafter\ifx\csname l@#1\endcsname\relax
\typeout{** WARNING: IEEEtran.bst: No hyphenation pattern has been}%
\typeout{** loaded for the language `#1'. Using the pattern for}%
\typeout{** the default language instead.}%
\else
\language=\csname l@#1\endcsname
\fi
#2}}
\providecommand{\BIBdecl}{\relax}
\BIBdecl

\bibitem{xiao2020motion}
X.~Xiao, B.~Liu, G.~Warnell, and P.~Stone, ``Motion control for mobile robot
  navigation using machine learning: a survey,'' \emph{arXiv preprint
  arXiv:2011.13112}, 2020.

\bibitem{liu2021lifelong}
B.~Liu, X.~Xiao, and P.~Stone, ``A lifelong learning approach to mobile robot
  navigation,'' \emph{IEEE Robotics and Automation Letters}, vol.~6, no.~2, pp.
  1090--1096, 2021.

\bibitem{xiao2021learning}
X.~Xiao, J.~Biswas, and P.~Stone, ``Learning inverse kinodynamics for accurate
  high-speed off-road navigation on unstructured terrain,'' in \emph{2021
  IEEE/RSJ International Conference on Intelligent Robots and Systems
  (IROS)}.\hskip 1em plus 0.5em minus 0.4em\relax IEEE, 2021.

\bibitem{wigness2018robot}
M.~Wigness, J.~G. Rogers, and L.~E. Navarro-Serment, ``Robot navigation from
  human demonstration: Learning control behaviors,'' in \emph{2018 IEEE
  International Conference on Robotics and Automation (ICRA)}.\hskip 1em plus
  0.5em minus 0.4em\relax IEEE, 2018, pp. 1150--1157.

\bibitem{richter2017safe}
C.~Richter and N.~Roy, ``Safe visual navigation via deep learning and novelty
  detection,'' 2017.

\bibitem{xiao2020appld}
X.~Xiao, B.~Liu, G.~Warnell, J.~Fink, and P.~Stone, ``Appld: Adaptive planner
  parameter learning from demonstration,'' \emph{IEEE Robotics and Automation
  Letters}, vol.~5, no.~3, pp. 4541--4547, 2020.

\bibitem{wang2021appli}
Z.~Wang, X.~Xiao, B.~Liu, G.~Warnell, and P.~Stone, ``Appli: Adaptive planner
  parameter learning from interventions,'' in \emph{2021 IEEE International
  Conference on Robotics and Automation (ICRA)}.\hskip 1em plus 0.5em minus
  0.4em\relax IEEE, 2021.

\bibitem{xu2021applr}
Z.~Xu, G.~Dhamankar, A.~Nair, X.~Xiao, G.~Warnell, B.~Liu, Z.~Wang, and
  P.~Stone, ``Applr: Adaptive planner parameter learning from reinforcement,''
  in \emph{2021 IEEE International Conference on Robotics and Automation
  (ICRA)}.\hskip 1em plus 0.5em minus 0.4em\relax IEEE, 2021.

\bibitem{xiao2021toward}
X.~Xiao, B.~Liu, G.~Warnell, and P.~Stone, ``Toward agile maneuvers in highly
  constrained spaces: Learning from hallucination,'' \emph{IEEE Robotics and
  Automation Letters}, pp. 1503--1510, 2021.

\bibitem{xiao2021agile}
X.~Xiao, B.~Liu, and P.~Stone, ``Agile robot navigation through hallucinated
  learning and sober deployment,'' in \emph{2021 IEEE International Conference
  on Robotics and Automation (ICRA)}.\hskip 1em plus 0.5em minus 0.4em\relax
  IEEE, 2021.

\bibitem{pfeiffer2017perception}
M.~Pfeiffer, M.~Schaeuble, J.~Nieto, R.~Siegwart, and C.~Cadena, ``From
  perception to decision: A data-driven approach to end-to-end motion planning
  for autonomous ground robots,'' in \emph{2017 ieee international conference
  on robotics and automation (icra)}.\hskip 1em plus 0.5em minus 0.4em\relax
  IEEE, 2017, pp. 1527--1533.

\bibitem{chiang2019learning}
H.-T.~L. Chiang, A.~Faust, M.~Fiser, and A.~Francis, ``Learning navigation
  behaviors end-to-end with autorl,'' \emph{IEEE Robotics and Automation
  Letters}, vol.~4, no.~2, pp. 2007--2014, 2019.

\bibitem{perille2020benchmarking}
D.~Perille, A.~Truong, X.~Xiao, and P.~Stone, ``Benchmarking metric ground
  navigation,'' in \emph{2020 IEEE International Symposium on Safety, Security,
  and Rescue Robotics (SSRR)}.\hskip 1em plus 0.5em minus 0.4em\relax IEEE,
  2020, pp. 116--121.

\bibitem{fox1997dynamic}
D.~Fox, W.~Burgard, and S.~Thrun, ``The dynamic window approach to collision
  avoidance,'' \emph{IEEE Robotics \& Automation Magazine}, vol.~4, no.~1, pp.
  23--33, 1997.

\bibitem{zhou2020ego}
X.~Zhou, Z.~Wang, H.~Ye, C.~Xu, and F.~Gao, ``Ego-planner: An esdf-free
  gradient-based local planner for quadrotors,'' \emph{IEEE Robotics and
  Automation Letters}, 2020.

\bibitem{lavalle2006planning}
S.~LaValle, \emph{Planning algorithms}.\hskip 1em plus 0.5em minus 0.4em\relax
  Cambridge university press, 2006.

\bibitem{chen2017socially}
Y.~F. Chen, M.~Everett, M.~Liu, and J.~P. How, ``Socially aware motion planning
  with deep reinforcement learning,'' in \emph{2017 IEEE/RSJ International
  Conference on Intelligent Robots and Systems (IROS)}.\hskip 1em plus 0.5em
  minus 0.4em\relax IEEE, 2017, pp. 1343--1350.

\bibitem{williams1992simple}
R.~J. Williams, ``Simple statistical gradient-following algorithms for
  connectionist reinforcement learning,'' \emph{Machine learning}, vol.~8,
  no.~3, pp. 229--256, 1992.

\bibitem{agrawal2019differentiable}
A.~Agrawal, B.~Amos, S.~Barratt, S.~Boyd, S.~Diamond, and Z.~Kolter,
  ``Differentiable convex optimization layers,'' \emph{arXiv preprint
  arXiv:1910.12430}, 2019.

\bibitem{anderson2018evaluation}
P.~Anderson, A.~Chang, D.~S. Chaplot, A.~Dosovitskiy, S.~Gupta, V.~Koltun,
  J.~Kosecka, J.~Malik, R.~Mottaghi, M.~Savva \emph{et~al.}, ``On evaluation of
  embodied navigation agents,'' \emph{arXiv preprint arXiv:1807.06757}, 2018.

\end{thebibliography}

\end{document}